\documentclass[10pt, a4paper]{article}

\usepackage{amsmath,amsfonts,bm}

\def\eqref#1{equation~\ref{#1}}

\def\1{\bm{1}}

\DeclareMathAlphabet{\mathsfit}{\encodingdefault}{\sfdefault}{m}{sl}
\SetMathAlphabet{\mathsfit}{bold}{\encodingdefault}{\sfdefault}{bx}{n}

\usepackage{mathtools}
\usepackage{multirow}
\usepackage{multicol}
\usepackage{booktabs}
\usepackage{subfigure}

\usepackage[T1]{fontenc}
\usepackage{subcaption}
\usepackage{url}
\usepackage{booktabs}       
\usepackage{amsfonts}       
\usepackage{nicefrac}       
\usepackage{microtype}      
\usepackage{xcolor} 
\usepackage{makecell}
\usepackage{colortbl} 
\usepackage{wrapfig}
\usepackage{makecell}
\usepackage{multirow}
\usepackage{graphicx}
\usepackage{caption}
\usepackage{float} 
\usepackage{pifont}
\usepackage{xcolor}
\usepackage{listings}

\definecolor{MutedGreen}{RGB}{85, 170, 85}
\definecolor{CoolAccent}{RGB}{120, 145, 230}
\definecolor{B}{RGB}{200,230,250} 
\definecolor{pink}{HTML}{F38181}

\usepackage{tcolorbox} 

\newtcolorbox{analysisbox}[1][]{
    colback=white,
    colframe=pink!75!black,
    fonttitle=\bfseries,
    boxsep=5pt,
    left=5pt,
    right=5pt,
    top=5pt,
    bottom=5pt,
    title=#1,
    width=\textwidth, 
}
\usepackage[final]{lrec2026} 

\title{Multi-Session Client-Centered Treatment Outcome Evaluation in Psychotherapy}

\name{Hongbin Na$^1$, Tao Shen$^1$, Shumao Yu$^2$, Ling Chen$^1$}
\address{$^1$Australian AI Institute, University of Technology Sydney, Australia \\
         $^2$KU Leuven, Belgium \\
         hongbin.na@student.uts.edu.au \\
        {\raisebox{-0.2em}{\includegraphics[height=1em]{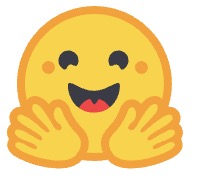}}}
\url{https://huggingface.co/datasets/UTSNLPGroup/TheraPhase}}

\abstract{
In psychotherapy, therapeutic outcome assessment, or treatment outcome evaluation, is essential to mental health care by systematically evaluating therapeutic processes and outcomes. Existing large language model approaches often focus on therapist-centered, single-session evaluations, neglecting the client's subjective experience and longitudinal progress across multiple sessions. To address these limitations, we propose IPAEval, a client-Informed Psychological Assessment-based Evaluation framework, which automates treatment outcome evaluations from the client's perspective using clinical interviews. It integrates cross-session client-contextual assessment and session-focused client-dynamics assessment for a comprehensive understanding of therapeutic progress. Specifically, IPAEval employs a two-stage prompt scheme that maps client information onto psychometric test items, enabling interpretable and structured psychological assessments. Experiments on our new TheraPhase dataset, comprising 400 paired initial and completion stage client records, demonstrate that IPAEval effectively tracks symptom severity and treatment outcomes over multiple sessions, outperforming baseline approaches across both closed-source and open-source models, and validating the benefits of items-aware reasoning mechanisms.
 \\ \newline  }

\begin{document}

\maketitleabstract

\section{Introduction} \label{sec:intro}
\begin{figure}[htbp]
    \centering
    \includegraphics[width=0.4\textwidth]{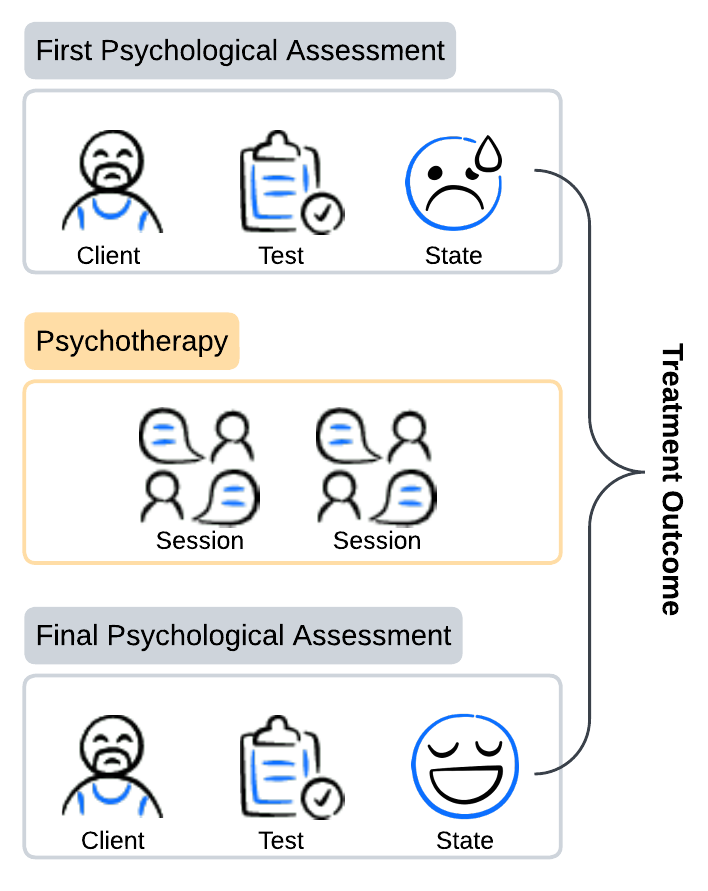}  
    \vspace{-5pt}
    \caption{What is Treatment Outcome?}
    \vspace{-10pt}
    \label{fig:treatmentoutcome}
\end{figure}

In psychotherapy, therapeutic outcome assessment, a.k.a treatment outcome (see Figure~\ref{fig:treatmentoutcome}) evaluation under clinical settings, refers to the systematic evaluation of therapeutic processes and outcomes~\citep{groth2009handbook},
focusing on factors such as therapist effectiveness~\citep{johns2019systematic} and treatment efficacy~\citep{jensen2018monitoring} to improve mental health care delivery. 
It plays a significant role in enhancing the quality and effectiveness of mental health care by providing actionable insights that guide therapists in refining their treatment approaches~\citep{wampold2015great}, ultimately leading to better client outcomes and improved therapeutic relationships in real-world clinical practice~\citep{Maruish2000TheUO}.

Over the last years, the emergence of large language models (LLMs) has demonstrated their effectiveness in automatic evaluations, showing a high degree of alignment with human judgment when provided with proper instruction and contextual guidance~\citep{liu-etal-2023-g, li2024leveraging, kim2024prometheus}.
This aligns with `LLMs-as-a-judge' paradigm, where LLMs are employed to simulate human evaluators by providing assessments upon natural language input~\citep{zheng2023judging, wang-etal-2024-large-language-models-fair}. 
This was extended to therapeutic outcome assessment by harnessing LLMs' ability to model complex therapeutic procedures and interactions, offering a novel pathway for automating the assessment of therapeutic efficacy~\citep{chiu2024computationalframeworkbehavioralassessment,lee2024cactuspsychologicalcounselingconversations,li2024automaticevaluationmentalhealth}.

\begin{table*}[t] 
\centering
\resizebox{\textwidth}{!}{%
\captionsetup{width=\textwidth}
\begin{tabular}{lcccc}
\toprule
\textbf{Method} & \textbf{Perspective} & \textbf{Theory Adherence} & \textbf{Reasoning} & \textbf{Evaluation Target} \\
\midrule
CPsyCoun~\citep{zhang2024cpsycoun}      & Therapist   & \textcolor{red}{\ding{55}} & \textcolor{red}{\ding{55}} & Single Session \\
Cactus~\citep{lee2024cactuspsychologicalcounselingconversations} & Therapist   &  \textcolor{green}{\ding{51}} & \textcolor{red}{\ding{55}} & Single Session \\
ClientCAST~\citep{wang2024clientcenteredassessmentllmtherapists} & Client      & \textcolor{green}{\ding{51}} & \textcolor{red}{\ding{55}} & Single Session \\
IPAEval (Ours)   & Client      & \textcolor{green}{\ding{51}} & \textcolor{green}{\ding{51}} & Multiple Sessions \\
\bottomrule
\end{tabular}
}
\vspace{-5pt}
\caption{Comparisons of IPAEval with other counterparts. 
\textbf{Perspective} indicates whether the evaluation is conducted from the therapist's or the client's point of view. \textbf{Theory Adherence} signifies whether the method is grounded in established psychological theories. \textbf{Reasoning} denotes whether the method involves generating intermediate reasoning steps before arriving at the final evaluation results. \textbf{Evaluation Target} refers to whether the method evaluates a single session or multiple sessions.}
\label{tab:method_comparison}
\vspace{-15pt}
\end{table*}

In the assessment, compared to psychometric tests~\citep{Furr_2020} that are often constrained by the limitations of self-reported data, susceptibility to social desirability biases~\citep{braun2001socially, paulhus2017socially}, 
clinical interviews not only provide richer, more nuanced insights into the client's emotional and behavioral states but also offer data that is more readily obtainable through natural, conversational interactions.
Therefore, many recent works leverage clinical interviews, potentially enriched by the client's profile~\citep{lee2024cactuspsychologicalcounselingconversations}, to evaluate therapists from multiple perspectives, including behavioral labels~\citep{chiu2024computationalframeworkbehavioralassessment}, skills adherence~\citep{lee2024cactuspsychologicalcounselingconversations}, and therapeutic rapport~\citep{li2024automaticevaluationmentalhealth, yosef-etal-2024-assessing}, offering a holistic view of their effectiveness in psychotherapy.

While the above therapist-centered assessments focus on evaluating the therapist's techniques and adherence to therapeutic models, they often overlook the subjective experience and evolving needs of the client, limiting the depth of the evaluation~\citep{wang2024clientcenteredassessmentllmtherapists, yosef-etal-2024-assessing}. In contrast, client-centered assessments, such as \textit{treatment outcome evaluation} in common practice, prioritize the client's perspective, offering a more comprehensive understanding of therapy's impact by capturing changes in the client's emotional, cognitive, and behavioral states across sessions~\citep{hatfield2004use, rogers2012client}. 
Although a concurrent work, ClientCAST \citep{wang2024clientcenteredassessmentllmtherapists}, presents an LLM-based client simulator for treatment outcome evaluations, which focuses on reducing harmful outputs and improving answering consistency, we stand fundamentally apart and never fabricate client responses that could distort the evaluation of treatment outcomes. 
What's worse, almost all previous approaches focus on evaluating individual therapy sessions in isolation, without considering the broader context of the client's journey across multiple sessions. 
This narrow scope limits the ability to assess longitudinal progress or capture the dynamic shifts in a client's mental state and therapeutic needs over time, which are crucial for a comprehensive treatment outcome evaluation~\citep{hayes2020complex}.

Motivated by the above therapist-centered and single-session limitations (please see Table~\ref{tab:method_comparison} for comparisons), we design a new evaluation framework, dubbed client-\textbf{I}nformed \textbf{P}sychological \textbf{A}ssessment-based \textbf{Eval}uation (IPAEval), for treatment outcomes in the format of clinical interviews. 

Specifically, to achieve treatment outcome evaluation, we formulate an information extraction task that leverages clinical interviews to automatically populate psychometric tests for psychological assessments, bridging the gap between subjective client dialogues and standardized metrics. As such, treatment outcomes are evaluated through these assessments of clients conducted both before and after therapy, allowing for a more comprehensive understanding of therapeutic progress. 
Upon this new framework, we first propose a cross-session client-contextual assessment module that integrates client history and contextual information across multiple sessions to enhance the accuracy of psychological assessments. Then, we present a session-focused client-dynamics assessment module that evaluates the effectiveness of individual therapy sessions by tracking real-time client responses and treatment outcomes within each session. 
In the meantime, to boost reasoning capability in the extraction, we also present an items-aware reasoning prompt technique for psychometric test-oriented rationale generation.

To evaluate the proposed framework, 
we first develop a new dataset, called TheraPhase, based on CPsyCoun~\citep{zhang2024cpsycoun}, which includes transcripts from initial and final therapy sessions. This dataset offers valuable insights into therapy progress and serves as a key resource for evaluating psychological assessments and treatment outcomes across multiple sessions.
Then, we tested nine LLMs, including both open- and closed-source models. These models were evaluated for their performance in psychological assessments and treatment outcome prediction, particularly in multi-session evaluations. IPAEval consistently tracked symptom severity and treatment outcomes across multiple sessions, a capability lacking in previous single-session models. Our ablation study confirmed that the items-aware reasoning mechanism significantly boosts model performance in both symptom detection and outcome prediction. 

\section{Methodology}

\begin{figure*}[t]
    \centering
    \includegraphics[width=1\linewidth]{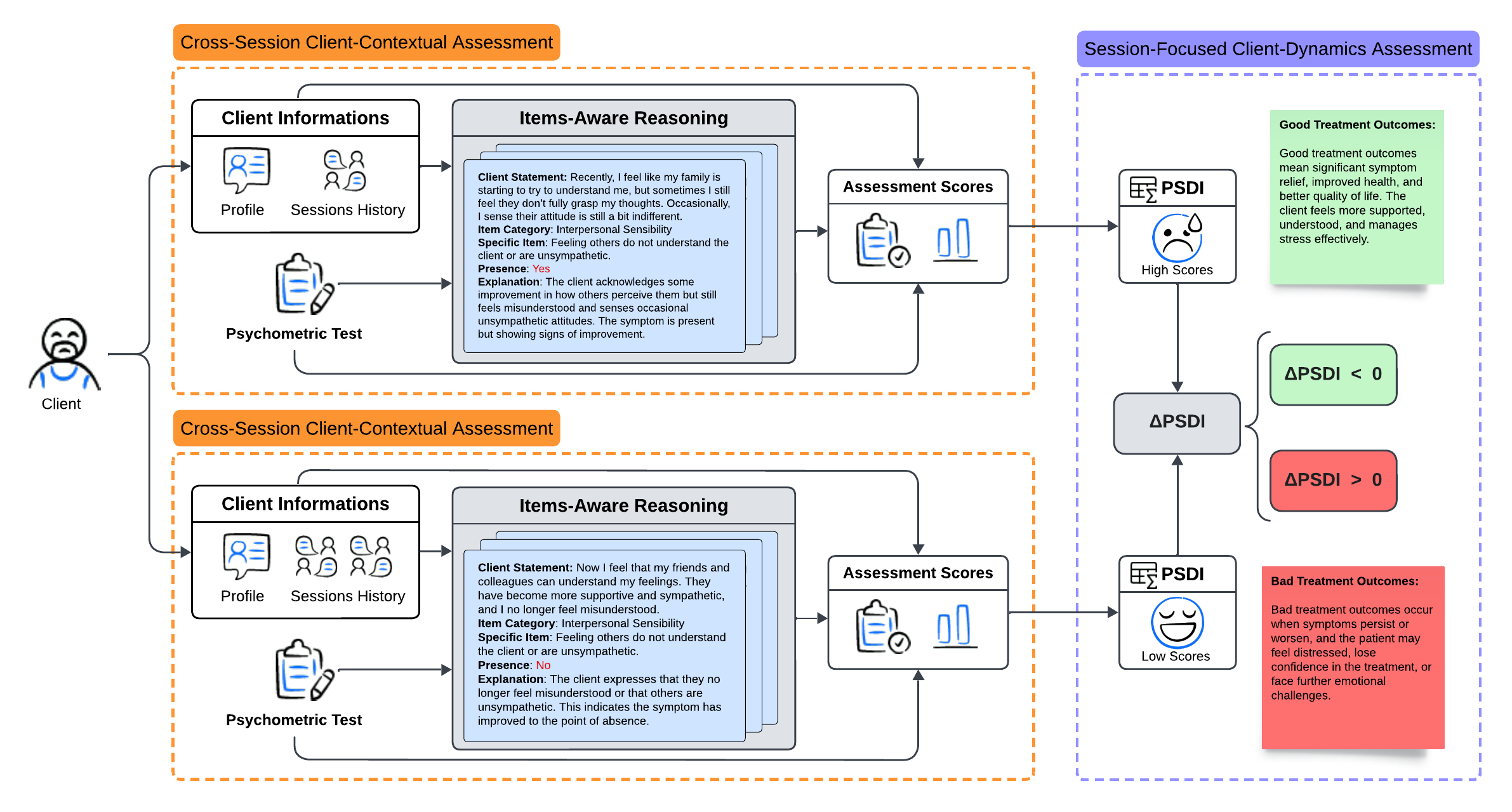}
    \vspace{-5pt}
    \caption{An illustration of client-informed psychological assessment-based evaluation (IPAEval).}
    \label{fig:overview}
    \vspace{-15pt}
\end{figure*}

Starting with a task definition (\S\ref{sec:task_def}), we elaborate on our evaluation framework, called client-Informed Psychological Assessment-based Evaluation (IPAEval), which is mainly composed of 1) a \textit{cross-session client-contextual assessment} module (\S\ref{sec:session-assess}) for client-tracking psychological assessment and 2) a \textit{session-focused client-dynamics assessment} module (\S\ref{sec:session-focus}) to derive session-informed treatment outcome evaluation. Please see Figure~\ref{fig:overview} for an overall illustration of our framework.
Last but not least, as there is no precursor in clinical interviews-based treatment outcome evaluation, we curate a new dataset, called TheraPhase, as a testbed for IPAEval (\S\ref{sec:TPD}).

\subsection{Task Definition} \label{sec:task_def}

Consider a client with profile $p$ undergoing a series of therapy sessions $s_1, s_2, \dots$. Our goal is to evaluate the treatment outcome of a given session $s_k$ (or a combination of consecutive sessions). We denote the client information available after session $s_k$ as $c_k = (p, s_k)$, which may be further enriched with prior session history when available (\S\ref{sec:session-assess}). We decompose the evaluation into two sequential sub-tasks, both performed by an LLM $\mathcal{M}$ instantiated differently for each stage.

\paragraph{Psychological Assessment.}
After session $s_k$, we conduct a psychological assessment based on the client information $c_k$ and a set of psychometric tests $\mathcal{T}$:
\begin{align}
    \mathbf{a}_k = \mathcal{M}^{(\text{a})}(c_k,\; \mathcal{T}),
    \label{eq:assess}
\end{align}
where $\mathbf{a}_k$ denotes the resulting assessment scores.

\paragraph{Treatment Outcome Evaluation.}
By comparing assessments from different stages, we quantify the treatment outcome:
\begin{align}
    \mathbf{e}_k = \mathcal{M}^{(\text{e})}(\mathbf{a}_k,\; \mathbf{a}_{<k}),
    \label{eq:outcome}
\end{align}
where $\mathbf{a}_{<k}$ refers to one or more prior assessments serving as the baseline. In the simplest case, $\mathbf{e}_k$ measures the change between an initial assessment and a post-treatment assessment. We detail the concrete instantiation in \S\ref{sec:session-focus}.

\subsection{Cross-session Client-contextual Assessment}\label{sec:session-assess}

Existing research using client information with LLMs for psychological assessment, particularly for depression and PTSD, shows promising results~\citep{11,23,28}. However, these studies typically focus on specific symptoms and lack broad coverage of psychological conditions and transparency in interpreting scale results, which may erode trust among clinicians and clients, limiting clinical applications~\citep{martin-rouas-2024-voice}.

To address these gaps, we introduce a two-stage prompt scheme that populates information from clinical interviews to fill psychometric tests by making the best of LLMs' capability in natural language understanding~\citep{zhao2023survey, hua2024applying}.
It is applicable to various psychometric tests and specifically designed to provide interpretable psychological assessments.
Without sacrificing generality, in this work we utilize a widely used and comprehensive psychometric test for screening psychological symptoms.

\paragraph{Stage 1: Items-Aware Reasoning.}
This stage extracts structured symptom evidence from client information. 
Inspired by \citet{schulhoff2024promptreportsystematicsurvey}, we design a prompt $p^{(\text{ir})}$ (\textbf{i}tems-aware \textbf{r}easoning) that instructs the LLM to act as a psychologist, map client information onto individual items of the psychometric test $\mathcal{T}$, and produce an explanation for each item. Concretely, the LLM generates items-aware reasoning results:
\begin{align}
    \hat{\mathcal{X}} = \underset{\mathcal{X}}{\text{argmax}}\; P_{\text{LLM}}(\mathcal{X} \mid c_k,\; p^{(\text{ir})}),
    \label{eq:items_reason}
\end{align}
where $c_k$ is the client information defined in \S\ref{sec:task_def} and $\hat{\mathcal{X}}$ is a set of structured records, each consisting of the extracted evidence, symptom category, specific symptom, presence judgment, and a detailed explanation.
This approach helps clinicians quickly trace the source of evidence and understand the relevance of various symptoms.
The detailed prompt and an example are provided in Appendix~\ref{appendix:promptIA} and Appendix~\ref{appendix:IAexample}, respectively.

\paragraph{Stage 2: Psychological Assessment.}
Building on the reasoning results, the LLM then scores the client across $N$ symptom dimensions. We design a second prompt $p^{(\text{sa})}$ (\textbf{s}ymptom \textbf{a}ssessment) that provides the psychometric test $\mathcal{T}$ together with simplified scoring criteria (at the dimension level rather than individual items, to account for the practical constraint that not all items are addressed in a session). The assessment scores are obtained as:
\begin{align}
    \hat{\mathbf{a}} = \underset{\mathbf{a}}{\text{argmax}}\; P_{\text{LLM}}(\mathbf{a} \mid c_k,\; \hat{\mathcal{X}},\; p^{(\text{sa})}),
    \label{eq:assess_score}
\end{align}
where $\hat{\mathbf{a}} \in \mathbb{R}^{N}$ contains the estimated score for each of the $N$ symptom dimensions.
The detailed prompt is provided in Appendix~\ref{appendix:promptPA}.

\paragraph{Remark: Avoiding Excessive Speculation.}
In contrast to ClientCAST~\citep{wang2024clientcenteredassessmentllmtherapists}, which simulates the client's own estimation of psychometric test scores, our approach adjusts score ranges to account for items not yet addressed by the client. This more accurately reflects the gradual disclosure of information over sessions and avoids biased assessments caused by unmentioned items.

\subsection{Session-focused Client-dynamics Assessment}\label{sec:session-focus}

Given the assessment scores $\hat{\mathbf{a}} \in \mathbb{R}^{N}$ over $N$ symptom dimensions, we compute the \textit{Positive Symptom Distress Index} (PSDI)~\citep{derogatis2010symptom}, which summarizes the average distress level across the dimensions where positive symptoms are detected. Let $\mathcal{P} \subseteq \{1,\dots,N\}$ denote the subset of dimensions with positive scores (i.e., $\hat{\mathbf{a}}_i > 0$). The PSDI is defined as:
\begin{align}
    \text{PSDI} = \frac{1}{|\mathcal{P}|} \sum_{i \in \mathcal{P}} \hat{\mathbf{a}}_i,
    \label{eq:PSDI}
\end{align}
where $|\mathcal{P}|$ denotes the number of positive dimensions.

To evaluate treatment outcomes, we apply the two-stage assessment (Eq.~\ref{eq:items_reason}--\ref{eq:assess_score}) to the client information at the initial stage $c_i$ and the final stage $c_f$, yielding $\text{PSDI}_i$ and $\text{PSDI}_f$ respectively. The treatment outcome is then defined as:
\begin{align}
    \mathbf{e} \coloneqq \Delta\text{PSDI} = \text{PSDI}_f - \text{PSDI}_i.
    \label{eq:Delta_PSDI}
\end{align}
A negative $\Delta\text{PSDI}$ indicates symptom improvement after treatment.

\paragraph{Remark: Advantages and Versatility of PSDI.}
Although PSDI originates from the Symptom Checklist-90 (SCL-90)~\citep{derogatis1973scl}, the idea of averaging scores over positive items generalizes naturally to other psychometric tests, providing a flexible tool for tracking treatment progress.

\subsection{Brief on TheraPhase Dataset Design} \label{sec:TPD}

Popular datasets such as High-Low Quality~\citep{perez-rosas-etal-2019-makes} and AnnoMI~\citep{fi15030110} contain only single-session client information sourced from public videos, with no data for subsequent stages. To assess cross-stage changes, we construct the TheraPhase Dataset based on CPsyCoun~\citep{zhang2024cpsycoun}, which exhibits significant within-session changes. Our dataset includes 400 pairs of client information from the initial and completion stages of treatment.

To construct the dataset, we use 5-shot prompting with GPT-4 to extract the initial-stage information from each client's comprehensive record, forming paired data (initial vs.\ full) that enables analytical comparison between pre- and post-treatment conditions. Please see \S\ref{sec:dataset_const} and \S\ref{appendix:datasets} for details.

\section{Experiments}

Starting with the settings of IPAEval framework and involved models, we elaborate on datasets constructions and their auto eval metrics (plus human alignment results), followed by empirical results, ablations, and error analysis.

\subsection{Experimental Settings}
\paragraph{IPAEval Setup.}The IPAEval framework is capable of handling various forms of client information, such as user profiles and interaction histories. However, due to data acquisition limitations, we primarily utilized consultation dialogue data as the main source of client information. 
Furthermore, IPAEval supports a variety of symptom-based psychometric tests, such as the General Health Questionnaire (GHQ) series~\citep{montazeri200312}, the Symptom Checklist (SCL) series, and the Brief Symptom Inventory (BSI)~\citep{derogatis1983brief}. 
In this experiment, we utilized the SCL-90~\citep{derogatis1973scl}, a widely recognized and comprehensive tool for assessing a broad range of psychological symptoms. The scoring criteria for assessing symptoms are outlined in Table~\ref{tab:sa}. Additionally, to ensure structured output, our code utilizes LangChain\footnote{\url{https://www.langchain.com/}} and Pydantic\footnote{\url{https://docs.pydantic.dev/}} for better LLMs integration and data validation.

\begin{table}[htbp]
\centering
\small
\setlength{\tabcolsep}{1.5pt}
\begin{tabular}{@{}c p{0.4\textwidth}@{}}
    \toprule
    \textbf{Score} & \textbf{Description} \\ 
    \midrule
    -1 & Symptom not addressed.\\ 
    0  & Symptom addressed, but no symptoms found; no signs of distress or dysfunction.\\ 
    1  & Minimal symptoms, minor indications of distress but no significant dysfunction.\\ 
    2  & Clear symptoms, clear indications of distress, and significant dysfunction.\\ 
    \bottomrule
\end{tabular}
\vspace{-5pt}
\caption{\small Scoring Criteria for Symptom Assessment}
\vspace{-10pt}
\label{tab:sa}
\end{table}

\paragraph{Involved Models.} We conducted an investigation into the performance of several closed and open-source LLMs. The closed-source models we tested include GPT-4~\citep{openai2024gpt4technicalreport}, GPT4o, GPT-4-turbo, and GPT-4o-mini, which represent the latest advancements in proprietary LLMs developed by OpenAI\footnote{Specific versions of the OpenAI models used in the tests were \texttt{gpt-4-0613}, \texttt{gpt-4o-2024-05-13}, \texttt{gpt-4-turbo-2024-04-09}, \texttt{gpt-4o-mini-2024-07-18}.}. Additionally, we tested a variety of open-source models, such as Llama3.1-405B~\citep{dubey2024llama3herdmodels}, Llama3.1-70B~\citep{dubey2024llama3herdmodels}, Qwen2-72B~\citep{yang2024qwen2technicalreport}, Mistral-8X22B~\citep{jiang2024mixtralexperts}, and Mistral-8X7B~\citep{jiang2024mixtralexperts}. These models vary significantly in terms of architecture, parameter size, and training data, providing a comprehensive overview of both commercial and community-driven LLM development. All of these models were invoked through API platforms\footnote{For the OpenAI models, we invoked them via \url{https://platform.openai.com}, Mistral models through \url{https://console.mistral.ai/}, Llama3.1 models via \url{https://fireworks.ai/}, and Qwen2 through \url{https://www.together.ai/}.}.

\subsection{Datasets Construction} \label{sec:dataset_const}
For \textbf{psychological assessment}, we selected 2 datasets, High-Low Quality Counseling~\citelanguageresource{HighLowCounseling_2019} and AnnoMI~\citelanguageresource{AnnoMI_2023}, consisting of counseling therapy transcripts extracted from publicly available videos on online platforms such as YouTube and Vimeo. But, there are issues of data duplication between these two datasets. Given the higher quality of data in AnnoMI, we have chosen to retain the AnnoMI data from the same sources. Furthermore, considering the context window limitation of one of our test models, GPT-4, the maximum number of dialogue turns is set to 102. To increase the challenge and ensure the dialogues are sufficiently complex for evaluating the model's capability in handling extended therapeutic conversations, the minimum number of turns is set at 25. Based on these, we have selected 110 client dialogue entries as our test data. 

For \textbf{treatment outcomes}, we selected the TheraPhase Dataset. This dataset comprises treatment session transcripts that encompass two distinct phases of client interactions. Its advantage lies in the clear changes observable in clients across these phases, which aids in observing the treatment outcomes. 
The statistics of the resulting datasets are listed in Appendix~\ref{appendix:datasets}. 

\subsection{Auto Evaluation and Human Alignment}
For \textbf{psychological assessment} (e.g., symptom detection), we assessed the model's ability to identify symptoms from client data using classification metrics such as Accuracy, Precision, Recall, and F1 scores (Binary, Macro, and Weighted). The scoring system assigned a value of $-1$ for a negative class and $0$, $1$, or $2$ for positive classes. For symptom severity, we computed a PSDI score for each client and reported error metrics -- Mean Squared Error (MSE) and Mean Absolute Error (MAE) -- across three runs to ensure consistency.

For \textbf{treatment outcomes}, we measured changes in positive symptom severity using $\Delta \text{PSDI}$, which reflects the difference in mean positive symptom scores between two assessments. A $\Delta \text{PSDI}$ greater than $0$ indicates worsening or new symptoms, while a value of $0$ or less suggests improvement or stability. We also evaluated the accuracy of predicting these changes using the same classification metrics.

 \paragraph{References Generation.} For psychological assessment, we manually annotated 30 client sessions with a Cohen's kappa of 0.73. Among the models evaluated, GPT-4o achieved the best performance, with an accuracy of 78.33\% and a binary F1 score of 72.95\%. Accordingly, GPT-4o was selected as the gold model for generating reference scores in this task. Similarly, for treatment outcomes evaluation, we annotated 60 sessions, reaching a Cohen's kappa of 0.81. GPT-4 delivered the top results with an accuracy of 74.44\% and a binary F1 score of 83.44\%, leading us to choose GPT-4 as the gold model for this evaluation. 
 For further experimental details regarding References Generation, please refer to the Appendix ~\ref{appendix:reference}.

\subsection{Evaluation Results across LLMs}

\begin{table*}[ht]
\centering
\resizebox{\textwidth}{!}{
\begin{tabular}{l|cccccc|cc}
\toprule[1.5pt]
\textbf{Models} & \textbf{Accuracy} $\uparrow$ & \textbf{Precision} $\uparrow$ & \textbf{Recall} $\uparrow$ & \textbf{F1}\textsubscript{Binary} $\uparrow$ & \textbf{F1}\textsubscript{Macro} $\uparrow$ & \textbf{F1}\textsubscript{Weighted} $\uparrow$ & \textbf{MSE} $\downarrow$ & \textbf{MAE} $\downarrow$ \\ 
\midrule
\rowcolor[gray]{0.88}
\multicolumn{9}{c}{\textit{Closed-Source Models}} \\ 
\midrule
GPT-4          & \cellcolor{B}0.7973{\scriptsize$\pm$0.01}      & 0.7852{\scriptsize$\pm$0.01}      & 0.7121{\scriptsize$\pm$0.01}      & \cellcolor{B}0.7469{\scriptsize$\pm$0.01}      & \cellcolor{B}0.7889{\scriptsize$\pm$0.01}      & \cellcolor{B}0.7956{\scriptsize$\pm$0.01}      & \cellcolor{B}0.2100{\scriptsize$\pm$0.02}      & \cellcolor{B}0.3292{\scriptsize$\pm$0.03}      \\
GPT-4-turbo    & 0.7561{\scriptsize$\pm$0.00}      & \cellcolor{B}0.8726{\scriptsize$\pm$0.02}      & 0.4913{\scriptsize$\pm$0.01}      & 0.6285{\scriptsize$\pm$0.01}      & 0.7234{\scriptsize$\pm$0.00}      & 0.7386{\scriptsize$\pm$0.00}      & 0.4055{\scriptsize$\pm$0.05}      & 0.4490{\scriptsize$\pm$0.03}      \\
GPT-4o-mini    & 0.4915{\scriptsize$\pm$0.00}      & 0.4467{\scriptsize$\pm$0.02}      & \cellcolor{B}0.8824{\scriptsize$\pm$0.01}      & 0.5931{\scriptsize$\pm$0.00}      & 0.4576{\scriptsize$\pm$0.01}      & 0.4359{\scriptsize$\pm$0.01}      & 0.2245{\scriptsize$\pm$0.01}      & 0.3329{\scriptsize$\pm$0.02}      \\ 
\midrule
\rowcolor[gray]{0.88}
\multicolumn{9}{c}{\textit{Open-Source Models}} \\ 
\midrule
Llama3.1-405B          & 0.7291{\scriptsize$\pm$0.00}      & 0.6960{\scriptsize$\pm$0.01}      & 0.6306{\scriptsize$\pm$0.00}      & \cellcolor{B}0.6616{\scriptsize$\pm$0.00}      & 0.7179{\scriptsize$\pm$0.00}      & 0.7269{\scriptsize$\pm$0.00}      & 0.3922{\scriptsize$\pm$0.03}      & 0.4476{\scriptsize$\pm$0.01}      \\
Qwen2-72B    & \cellcolor{B}0.7385{\scriptsize$\pm$0.00}      & \cellcolor{B}0.7405{\scriptsize$\pm$0.01}      & 0.5815{\scriptsize$\pm$0.01}      & 0.6513{\scriptsize$\pm$0.01}      & \cellcolor{B}0.7210{\scriptsize$\pm$0.00}      & \cellcolor{B}0.7322{\scriptsize$\pm$0.00}      & 0.3962{\scriptsize$\pm$0.01}      & 0.4559{\scriptsize$\pm$0.00}      \\
Llama3.1-70B    & 0.7333{\scriptsize$\pm$0.01}      & 0.7201{\scriptsize$\pm$0.01}      & 0.5974{\scriptsize$\pm$0.01}      & 0.6529{\scriptsize$\pm$0.01}      & 0.7182{\scriptsize$\pm$0.01}      & 0.7286{\scriptsize$\pm$0.01}      & \cellcolor{B}0.3379{\scriptsize$\pm$0.01}      & \cellcolor{B}0.4041{\scriptsize$\pm$0.00}      \\
Mistral-8X22B    & 0.6215{\scriptsize$\pm$0.00}      & 0.5405{\scriptsize$\pm$0.00}      & \cellcolor{B}0.6616{\scriptsize$\pm$0.02}      & 0.5948{\scriptsize$\pm$0.00}      & 0.6198{\scriptsize$\pm$0.00}      & 0.6238{\scriptsize$\pm$0.00}      & 0.5205{\scriptsize$\pm$0.03}      & 0.5452{\scriptsize$\pm$0.02}      \\
Mistral-8X7B    & 0.6070{\scriptsize$\pm$0.00}      & 0.6158{\scriptsize$\pm$0.01}      & 0.1710{\scriptsize$\pm$0.02}      & 0.2672{\scriptsize$\pm$0.02}      & 0.4993{\scriptsize$\pm$0.01}      & 0.5364{\scriptsize$\pm$0.01}      & 1.5711{\scriptsize$\pm$0.02}      & 1.0927{\scriptsize$\pm$0.01}      \\
\bottomrule
\end{tabular}
}
\vspace{-5pt}
\caption{Performance comparison between closed-source and open-source models across various evaluation metrics in \textbf{psychological assessment}. Metrics with an upward arrow $\uparrow$ indicate higher values are better, while metrics with a downward arrow $\downarrow$ indicate lower values are better. The results show mean values along with standard deviations for each metric. Cells highlighted in \colorbox{B}{blue} represent the best-performing results.}
\vspace{-5pt}
\label{tab:RQ1M}
\end{table*}
\begin{table*}[h]
\centering
\resizebox{0.8\textwidth}{!}{
\begin{tabular}{l|cccccc}
\toprule[1.5pt]
\textbf{Models} & \textbf{Accuracy} $\uparrow$ & \textbf{Precision} $\uparrow$ & \textbf{Recall} $\uparrow$ & \textbf{F1}\textsubscript{Binary} $\uparrow$ & \textbf{F1}\textsubscript{Macro} $\uparrow$ & \textbf{F1}\textsubscript{Weighted} $\uparrow$ \\ 
\midrule
\rowcolor[gray]{0.88}
\multicolumn{7}{c}{\textit{Closed-Source Models}} \\ 
\midrule
GPT-4o          & 0.6375{\scriptsize$\pm$0.02}      & 0.7706{\scriptsize$\pm$0.01}      & 0.7356{\scriptsize$\pm$0.02}      & 0.7526{\scriptsize$\pm$0.01}      & 0.5370{\scriptsize$\pm$0.02}      & 0.6448{\scriptsize$\pm$0.02}      \\
GPT-4-turbo    & \cellcolor{B}0.6800{\scriptsize$\pm$0.01}      & \cellcolor{B}0.7824{\scriptsize$\pm$0.01}      & \cellcolor{B}0.7944{\scriptsize$\pm$0.01}      & \cellcolor{B}0.7883{\scriptsize$\pm$0.00}      & \cellcolor{B}0.5660{\scriptsize$\pm$0.01}      & \cellcolor{B}0.6772{\scriptsize$\pm$0.01}      \\
GPT-4o-mini    & 0.6317{\scriptsize$\pm$0.01}      & 0.7727{\scriptsize$\pm$0.01}      & 0.7211{\scriptsize$\pm$0.01}      & 0.7459{\scriptsize$\pm$0.01}      & 0.5380{\scriptsize$\pm$0.01}      & 0.6420{\scriptsize$\pm$0.01}      \\ 
\midrule
\rowcolor[gray]{0.88}
\multicolumn{7}{c}{\textit{Open-Source Models}} \\ 
\midrule
Llama3.1-405B          & \cellcolor{B}0.6958{\scriptsize$\pm$0.01}      & \cellcolor{B}0.7965{\scriptsize$\pm$0.01}      & 0.7989{\scriptsize$\pm$0.01}      & 0.7976{\scriptsize$\pm$0.00}      & \cellcolor{B}0.5925{\scriptsize$\pm$0.02}      & \cellcolor{B}0.6951{\scriptsize$\pm$0.01}      \\
Qwen2-72B    & 0.6725{\scriptsize$\pm$0.01}      & 0.7747{\scriptsize$\pm$0.01}      & 0.7944{\scriptsize$\pm$0.01}      & 0.7844{\scriptsize$\pm$0.01}      & 0.5515{\scriptsize$\pm$0.01}      & 0.6679{\scriptsize$\pm$0.01}      \\
Llama3.1-70B    & 0.6708{\scriptsize$\pm$0.01}      & 0.7796{\scriptsize$\pm$0.01}      & 0.7822{\scriptsize$\pm$0.01}      & 0.7809{\scriptsize$\pm$0.01}      & 0.5597{\scriptsize$\pm$0.02}      & 0.6703{\scriptsize$\pm$0.01}      \\
Mistral-8X22B    & 0.6383{\scriptsize$\pm$0.01}      & 0.7544{\scriptsize$\pm$0.00}      & 0.7678{\scriptsize$\pm$0.01}      & 0.7610{\scriptsize$\pm$0.01}      & 0.5089{\scriptsize$\pm$0.01}      & 0.6350{\scriptsize$\pm$0.01}      \\
Mistral-8X7B    & 0.6825{\scriptsize$\pm$0.01}      & 0.7469{\scriptsize$\pm$0.00}      & \cellcolor{B}0.8722{\scriptsize$\pm$0.02}      & \cellcolor{B}0.8046{\scriptsize$\pm$0.01}      & 0.4779{\scriptsize$\pm$0.00}      & 0.6413{\scriptsize$\pm$0.00}      \\
\bottomrule
\end{tabular}
}
\vspace{-5pt}
\caption{Comparison between closed and open-source models across various evaluation metrics in \textbf{treatment outcomes}.}
\vspace{-15pt}
\label{tab:RQ2M}
\end{table*}

\paragraph{Psychological Assessments.} As shown in Table~\ref{tab:RQ1M}, GPT-4 achieved the best performance in symptom detection, excelling in both accuracy and binary F1 score, highlighting its strong ability to accurately identify symptoms. GPT-4-turbo demonstrated a more conservative approach with higher precision but lower recall, indicating it was more cautious in detecting symptoms but missed more cases. GPT-4o-mini excelled in recall but had reduced overall reliability due to a higher rate of false positives. Among open-source models, Qwen2-72B and Llama3.1-70B showed the closest performance to GPT-4, though they still fell short. Notably, Mistral-8X7B's extremely low recall was caused by a significant number of output formatting errors, leading to evaluation failures. We will further discuss these formatting issues in Appendix~\ref{app:errors}.

In symptom severity assessment, GPT-4 once again stood out with the lowest MSE and MAE, making it the most accurate model. Although GPT-4o-mini and GPT-4-turbo showed more balanced results, they were less precise compared to GPT-4. Among open-source models, Llama3.1-70B performed the best, though the gap between open-source and closed-source models remained substantial. Furthermore, GPT-4 exhibited the greatest consistency and reliability, with minimal variance across runs, indicating robust performance. In contrast, GPT-4o-mini showed more variability in MAE and MSE, and open-source models generally exhibited less stability compared to their closed-source counterparts.

\paragraph{Treatment Outcomes.} Table~\ref{tab:RQ2M} compares the performance of closed-source and open-source models on treatment outcome evaluation tasks. Among the closed-source models, GPT-4-turbo achieved the highest scores across multiple metrics, making it the most effective model in treatment outcome prediction. GPT-4o and GPT-4o-mini displayed competitive performance but lagged slightly behind GPT-4-turbo. For the open-source models, Llama3.1-405B led the group with the highest accuracy and macro F1, demonstrating superior performance in treatment outcome tasks. Qwen2-72B and Llama3.1-70B also performed well, while Mistral-8X7B had the highest recall but struggled with lower F1 scores, indicating higher sensitivity but less consistent overall performance. Overall, both closed-source and open-source models showed strong capabilities, with GPT-4-turbo and Llama3.1-405B emerging as the top performers in their respective categories.

\subsection{Further Analysis}
\begin{figure*}[t]
    \centering
    \includegraphics[width=1\linewidth]{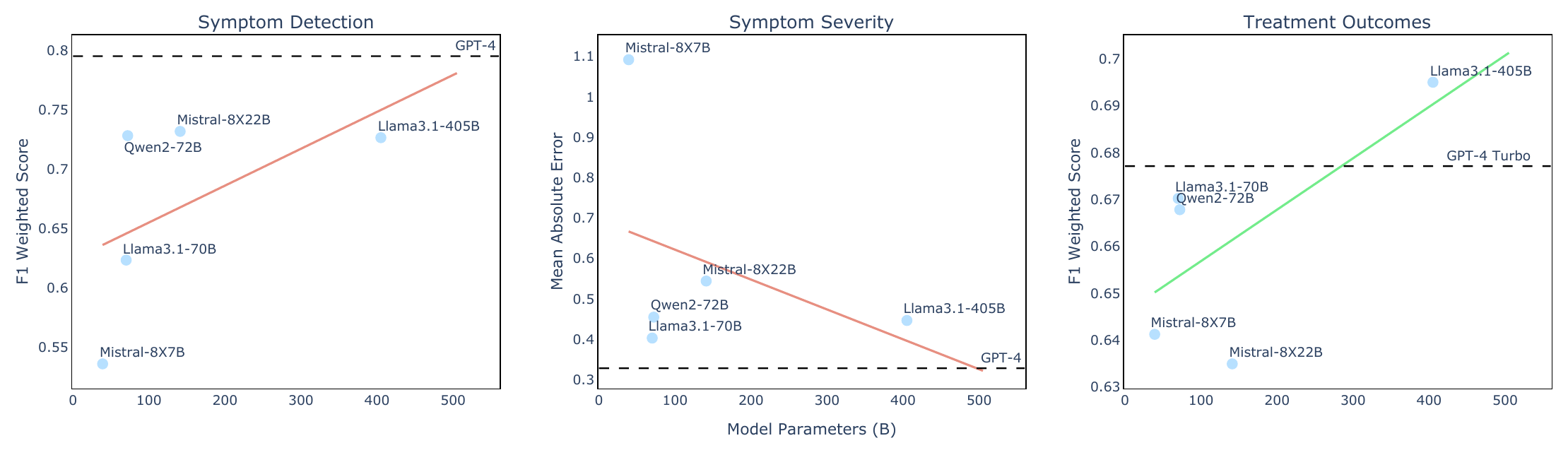}
    \vspace{-15pt}
    \caption{The impact of model parameters on symptom detection, symptom severity evaluation, and treatment outcome prediction. Dashed lines represent the best-performing closed-source models.}
    \label{fig:scale}
    \vspace{-12pt}
\end{figure*}
\begin{figure*}[t]
    \centering
    \includegraphics[width=1\linewidth]{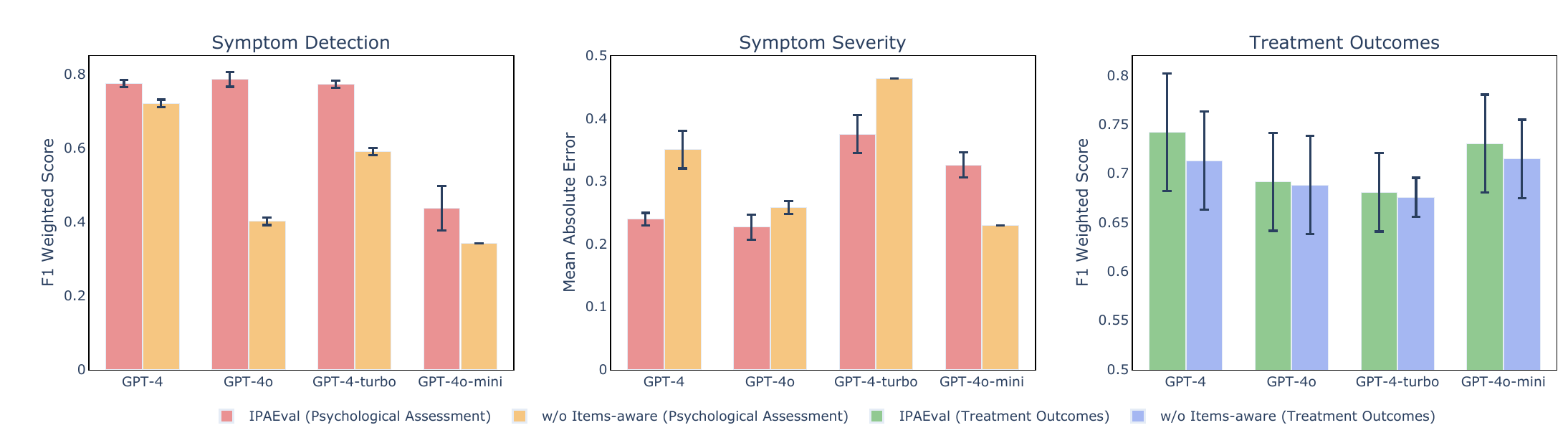}
    \vspace{-15pt}
    \caption{The impact of items-aware reasoning on psychological assessment and treatment outcomes evaluation using human-annotated data across four OpenAI models.}
    \label{fig:ablation}
    \vspace{-12pt}
\end{figure*}

\paragraph{Impact of Parameters on Performance.}
As shown in Figure~\ref{fig:scale}, model parameter size has a clear impact on performance across tasks such as symptom detection, symptom severity evaluation, and treatment outcome prediction. Larger models consistently outperform smaller models, exhibiting higher F1 (Weighted) scores and lower MAE. This trend indicates that increasing model size enhances the model's ability to handle complex tasks~\citep{wen2024benchmarking}, especially in identifying subtle patterns related to psychological symptoms and predicting treatment outcomes.

\paragraph{Impact of Items-aware Reasoning.} The ablation study (Figure~\ref{fig:ablation}) demonstrates that items-aware reasoning is crucial for both psychological assessment and treatment outcomes evaluation. Removing this feature significantly decreased performance across all models. For psychological assessment, models like GPT-4o and GPT-4 less accurately detected symptoms and assessed severity, which lowered F1 scores and increased error metrics. Similarly, performance in treatment outcomes evaluation also dropped, though the impact was less pronounced. These results underscore that items-aware reasoning improves the models' precision for these tasks.


\section{Related Work}

\paragraph{Therapist Assessment using LLMs.}
LLMs' role-playing capabilities have led to increased interest in developing Role-Play Therapists~\citep{chen2023llmempoweredchatbotspsychiatristpatient,chiu2024computationalframeworkbehavioralassessment,lee2024cactuspsychologicalcounselingconversations}, but the lack of automated metrics for evaluating t
herapist is a significant challenge.
CPsyCoun~\citep{zhang2024cpsycoun} employs an LLM-based evaluation method from the therapist's perspective to assess single session, specifically evaluating the therapist's comprehensiveness, professionalism, authenticity, and safety.
\citet{lee2024cactuspsychologicalcounselingconversations},~\citet{li2024automaticevaluationmentalhealth}, and~\citet{yosef-etal-2024-assessing} similarly adopt a therapist's perspective with LLM-based evaluation, but they address CPsyCoun's lack of support from psychological theories by employing the Cognitive Therapy Rating Scale~\citep{goldberg2020structure} for CBT skills assessment and the Working Alliance Inventory~\citep{hatcher2006development} for evaluating the therapeutic relationship. Notably, BOLT~\citep{chiu2024computationalframeworkbehavioralassessment} applied LLMs to identify therapist behaviors, evaluating the quality of dialogue sessions based on the frequency and sequence of LLM therapist behaviors. Clinical evidence~\citep{goodson2017impact, mason2016my} shows that better therapists are linked to improved outcomes, but evaluating therapists alone may miss how much the client is benefiting~\citep{robinson2009therapist}. The treatment outcome evaluation based on client-centered psychological assessment focuses more on results, specifically determining whether the therapy has brought about meaningful changes in the client's life, which is the ultimate goal of the treatment~\citep{groth2009handbook}.

\paragraph{Client-centered Psychological Assessment. }
Client-centered psychological assessment combines psychometric tests with clinical interviews to provide a comprehensive understanding of individuals~\citep{spoto2013theoretical}. While tests offer standardized data on psychological traits, interviews yield deeper insights into personal experiences, addressing nuances that tests might miss~\citep{groth2009handbook}. Using multiple methods ensures a complete client assessment in clinical practice~\citep{meyer2001psychological, groth2009handbook}. Moreover, leveraging LLMs' advanced language processing capabilities~\citep{luo2023chatgptfactualinconsistencyevaluator, zhao2023surveylargelanguagemodels} enables complex and diverse assessments, contrasting with earlier approaches that only detected individual symptoms~\citep{ji-etal-2022-mentalbert, zhai2024chinese}. A substantial body of research supports this advancement~\citep{11,23,28}. For example, several studies have employed LLMs to analyze interviews~\citep{gratch-etal-2014-distress}, assessing depression and PTSD scores via widely used tests like~\citep{kroenke2009phq} and PCL-C~\citep{weathers1994ptsd}.
However, precise psychological assessments enable therapists to grasp the client's psychological state, but a psychological assessment alone cannot determine whether the treatment has brought about positive changes for the client. 

Treatment outcomes evaluation complements psychological assessment by measuring the effectiveness of interventions over time~\citep{Maruish2000TheUO}. While psychological assessments provide a snapshot of the client's mental state, as shown in the Figure~\ref{fig:treatmentoutcome}, treatment outcomes evaluation focuses on tracking changes in symptoms and overall well-being throughout the therapeutic process. This dynamic evaluation allows therapists to determine whether the treatment has been successful and as needed for improvement.

\section{Conclusion}
We presented IPAEval to overcome limitations of current therapeutic outcome evaluations by shifting the focus from therapist-centered, single-session assessments to a comprehensive, client-informed framework. By using clinical interviews and integrating cross-session client-contextual and session-focused client-dynamics assessments, it offers a holistic evaluation of treatment outcomes. Experiments on our TheraPhase dataset validate its effectiveness in tracking symptom severity and progress across sessions. 
Overall, our results underscore the value of a client-centered, multi-session evaluation strategy for personalized and effective mental health interventions.

\section*{Limitations}
The limitations of this paper are as follows: (1) Due to the shortage of professional psychological annotators, only two individuals were involved in a limited amount of data labeling. This resulted in fewer human-aligned experimental data. Future research should focus on developing more multi-session datasets that include psychological assessment scores. (2) As the amount of client information increases, smaller models with fewer parameters struggle to follow instructions effectively. This limits the scalability and performance of these models in more complex scenarios. Future research should explore strategies to enhance model adaptability in handling larger client information inputs. (3) Our multi-session dataset was derived by splitting single multi-turn conversations, which can represent client changes but cannot fully capture the characteristics of true multi-session data. In the future, we should develop authentic multi-session datasets.

\section*{Ethics Statement}\label{sec:ethics}
All datasets used in this study are either fully synthetic—thus free of personally identifiable information—or have been rigorously anonymized and are employed strictly under their original usage agreements, eliminating privacy concerns. Nonetheless, large language models can generate inaccurate or biased outputs; relying on these results without professional oversight could mislead or harm clients. Consequently, IPAEval is intended solely as a \emph{research-level} evaluation tool, never a substitute for clinical diagnosis or treatment, and every automated conclusion must be reviewed by a licensed mental-health professional. Because psychological symptoms manifest differently across cultures, genders, and age groups—and our current training data are primarily English- and Chinese-centric—we recognize limitations in cross-cultural generalization. 

\section*{Bibliographical References}\label{sec:reference}
\bibliographystyle{lrec2026-natbib}
\bibliography{lrec2026-example}

@dataset{AnnoMI_2023,
  author       = {Wu, Zixiu and Balloccu, Simone and Kumar, Vivek and Helaoui, Rim and Reforgiato Recupero, Diego and Riboni, Daniele},
  title        = {AnnoMI: Expert-Annotated Counselling Dialogues},
  year         = {2023},
  version      = {Full release},
  howpublished = {Dataset},
  publisher    = {University of Aberdeen & University of Cagliari},
  doi          = {10.3390/fi15030110},
  url          = {https://github.com/uccollab/AnnoMI},
  note         = {Professionally transcribed MI dialogues with expert annotations; openly available.}
}

@dataset{HighLowCounseling_2019,
  author       = {P{\'e}rez-Rosas, Ver{\'o}nica and Wu, Xinyi and Resnicow, Kenneth and Mihalcea, Rada},
  title        = {High- vs. Low-Quality Counseling Conversations Corpus},
  year         = {2019},
  howpublished = {Dataset derived from public video sources},
  publisher    = {Associated with ACL 2019 paper},
  url          = {https://aclanthology.org/P19-1088/},
  note         = {Transcripts from YouTube/Vimeo demonstration videos; availability varies; see paper for details.}
}

@book{groth2009handbook,
  title={Handbook of psychological assessment},
  author={Groth-Marnat, Gary},
  year={2009},
  publisher={John Wiley \& Sons}
}

@article{derogatis1973scl,
  title={SCL-90: an outpatient psychiatric rating scale--preliminary report},
  author={Derogatis, Leonard R and Lipman, Ronald S and Covi, Lino},
  journal={Psychopharmacol bull},
  volume={9},
  number={1},
  pages={13--28},
  year={1973}
}

@article{jensen2018monitoring,
  title={Monitoring treatment progress and providing feedback is viewed favorably but rarely used in practice},
  author={Jensen-Doss, Amanda and Haimes, Emily M Becker and Smith, Ashley M and Lyon, Aaron R and Lewis, Cara C and Stanick, Cameo F and Hawley, Kristin M},
  journal={Administration and Policy in Mental Health and Mental Health Services Research},
  volume={45},
  pages={48--61},
  year={2018},
  publisher={Springer}
}

@book{wampold2015great,
  title={The great psychotherapy debate: The evidence for what makes psychotherapy work},
  author={Wampold, Bruce E and Imel, Zac E},
  year={2015},
  publisher={Routledge}
}

@article{Maruish2000TheUO,
  title={The Use of Psychological Testing for Treatment Planning and Outcome Assessment},
  author={Mark E. Maruish and Robert L. Leahy},
  journal={Journal of Cognitive Psychotherapy},
  year={2000},
  volume={14},
  pages={205 - 206}
}

@inbook{Furr_2020, 
    place={Cambridge}, 
    series={Cambridge Handbooks in Psychology}, 
    title={Psychometrics in Clinical Psychological Research}, 
    booktitle={The Cambridge Handbook of Research Methods in Clinical Psychology},
    publisher={Cambridge University Press}, 
    author={Furr, R. Michael},
    editor={Wright, Aidan G. C. and Hallquist, Michael N.Editors},
    year={2020}, 
    pages={54–65},
    collection={Cambridge Handbooks in Psychology}
}

@incollection{braun2001socially,
  title={Socially desirable responding: The evolution of a construct},
  author={Braun, Henry I and Jackson, Douglas N and Wiley, David E},
  booktitle={The role of constructs in psychological and educational measurement},
  pages={61--84},
  year={2001},
  publisher={Routledge}
}

@article{paulhus2017socially,
  title={Socially desirable responding on self-reports},
  author={Paulhus, Delroy L},
  journal={Encyclopedia of personality and individual differences},
  volume={1},
  number={5},
  year={2017},
  publisher={Springer Cham, Switzerland}
}

@misc{li2024leveraging,
    title={Leveraging Large Language Models for NLG Evaluation: Advances and Challenges},
    author={Zhen Li and Xiaohan Xu and Tao Shen and Can Xu and Jia-Chen Gu and Yuxuan Lai and Chongyang Tao and Shuai Ma},
    year={2024},
    eprint={2401.07103},
    archivePrefix={arXiv},
    primaryClass={cs.CL}
}

@misc{kim2024prometheus,
      title={Prometheus 2: An Open Source Language Model Specialized in Evaluating Other Language Models}, 
      author={Seungone Kim and Juyoung Suk and Shayne Longpre and Bill Yuchen Lin and Jamin Shin and Sean Welleck and Graham Neubig and Moontae Lee and Kyungjae Lee and Minjoon Seo},
      year={2024},
      eprint={2405.01535},
      archivePrefix={arXiv},
      primaryClass={cs.CL}
}

@misc{zhao2023survey,
    title={A Survey of Large Language Models},
    author={Wayne Xin Zhao and Kun Zhou and Junyi Li and Tianyi Tang and Xiaolei Wang and Yupeng Hou and Yingqian Min and Beichen Zhang and Junjie Zhang and Zican Dong and Yifan Du and Chen Yang and Yushuo Chen and Zhipeng Chen and Jinhao Jiang and Ruiyang Ren and Yifan Li and Xinyu Tang and Zikang Liu and Peiyu Liu and Jian-Yun Nie and Ji-Rong Wen},
    year={2023},
    eprint={2303.18223},
    archivePrefix={arXiv},
    primaryClass={cs.CL}
}

@misc{lee2024cactuspsychologicalcounselingconversations,
      title={Cactus: Towards Psychological Counseling Conversations using Cognitive Behavioral Theory}, 
      author={Suyeon Lee and Sunghwan Kim and Minju Kim and Dongjin Kang and Dongil Yang and Harim Kim and Minseok Kang and Dayi Jung and Min Hee Kim and Seungbeen Lee and Kyoung-Mee Chung and Youngjae Yu and Dongha Lee and Jinyoung Yeo},
      year={2024},
      eprint={2407.03103},
      archivePrefix={arXiv},
      primaryClass={cs.CL},
      url={https://arxiv.org/abs/2407.03103}, 
}

@misc{li2024automaticevaluationmentalhealth,
      title={Automatic Evaluation for Mental Health Counseling using LLMs}, 
      author={Anqi Li and Yu Lu and Nirui Song and Shuai Zhang and Lizhi Ma and Zhenzhong Lan},
      year={2024},
      eprint={2402.11958},
      archivePrefix={arXiv},
      primaryClass={cs.CL},
      url={https://arxiv.org/abs/2402.11958}, 
}

@misc{chiu2024computationalframeworkbehavioralassessment,
      title={A Computational Framework for Behavioral Assessment of LLM Therapists}, 
      author={Yu Ying Chiu and Ashish Sharma and Inna Wanyin Lin and Tim Althoff},
      year={2024},
      eprint={2401.00820},
      archivePrefix={arXiv},
      primaryClass={cs.CL},
      url={https://arxiv.org/abs/2401.00820}, 
}

@inproceedings{yosef-etal-2024-assessing,
    title = "Assessing Motivational Interviewing Sessions with {AI}-Generated Patient Simulations",
    author = "Yosef, Stav  and
      Zisquit, Moreah  and
      Cohen, Ben  and
      Klomek Brunstein, Anat  and
      Bar, Kfir  and
      Friedman, Doron",
    editor = "Yates, Andrew  and
      Desmet, Bart  and
      Prud{'}hommeaux, Emily  and
      Zirikly, Ayah  and
      Bedrick, Steven  and
      MacAvaney, Sean  and
      Bar, Kfir  and
      Ireland, Molly  and
      Ophir, Yaakov",
    booktitle = "Proceedings of the 9th Workshop on Computational Linguistics and Clinical Psychology (CLPsych 2024)",
    month = mar,
    year = "2024",
    address = "St. Julians, Malta",
    publisher = "Association for Computational Linguistics",
    url = "https://aclanthology.org/2024.clpsych-1.1",
    pages = "1--11",
}

@inproceedings{martin-rouas-2024-voice,
    title = "Why Voice Biomarkers of Psychiatric Disorders Are Not Used in Clinical Practice? Deconstructing the Myth of the Need for Objective Diagnosis",
    author = "Martin, Vincent P.  and
      Rouas, Jean-Luc",
    editor = "Calzolari, Nicoletta  and
      Kan, Min-Yen  and
      Hoste, Veronique  and
      Lenci, Alessandro  and
      Sakti, Sakriani  and
      Xue, Nianwen",
    booktitle = "Proceedings of the 2024 Joint International Conference on Computational Linguistics, Language Resources and Evaluation (LREC-COLING 2024)",
    month = may,
    year = "2024",
    address = "Torino, Italia",
    publisher = "ELRA and ICCL",
    url = "https://aclanthology.org/2024.lrec-main.1531",
    pages = "17603--17613",
}

@misc{zhang2024cpsycoun,
      title={CPsyCoun: A Report-based Multi-turn Dialogue Reconstruction and Evaluation Framework for Chinese Psychological Counseling}, 
      author={Chenhao Zhang and Renhao Li and Minghuan Tan and Min Yang and Jingwei Zhu and Di Yang and Jiahao Zhao and Guancheng Ye and Chengming Li and Xiping Hu},
      year={2024},
      eprint={2405.16433},
      archivePrefix={arXiv},
      primaryClass={cs.CL}
}

@misc{schulhoff2024promptreportsystematicsurvey,
      title={The Prompt Report: A Systematic Survey of Prompting Techniques}, 
      author={Sander Schulhoff and Michael Ilie and Nishant Balepur and Konstantine Kahadze and Amanda Liu and Chenglei Si and Yinheng Li and Aayush Gupta and HyoJung Han and Sevien Schulhoff and Pranav Sandeep Dulepet and Saurav Vidyadhara and Dayeon Ki and Sweta Agrawal and Chau Pham and Gerson Kroiz and Feileen Li and Hudson Tao and Ashay Srivastava and Hevander Da Costa and Saloni Gupta and Megan L. Rogers and Inna Goncearenco and Giuseppe Sarli and Igor Galynker and Denis Peskoff and Marine Carpuat and Jules White and Shyamal Anadkat and Alexander Hoyle and Philip Resnik},
      year={2024},
      eprint={2406.06608},
      archivePrefix={arXiv},
      primaryClass={cs.CL},
      url={https://arxiv.org/abs/2406.06608}, 
}

@misc{11,  
title={The Capability of Large Language Models to Measure Psychiatric Functioning},  
author={Isaac R. Galatzer-Levy and Daniel McDuff and Vivek Natarajan and Alan Karthikesalingam and Matteo Malgaroli},  
year={2023},  
eprint={2308.01834},  
archivePrefix={arXiv},  
primaryClass={[cs.CL](http://cs.CL)},  
url={[https://arxiv.org/abs/2308.01834](https://arxiv.org/abs/2308.01834)},

}

@misc{23,  
title={An Assessment on Comprehending Mental Health through Large Language Models},  
author={Mihael Arcan and David-Paul Niland and Fionn Delahunty},  
year={2024},  
eprint={2401.04592},  
archivePrefix={arXiv},  
primaryClass={[cs.CL](http://cs.CL)},  
url={[https://arxiv.org/abs/2401.04592](https://arxiv.org/abs/2401.04592)},

}

@misc{28,  
title={LLM Questionnaire Completion for Automatic Psychiatric Assessment},  
author={Gony Rosenman and Lior Wolf and Talma Hendler},  
year={2024},  
eprint={2406.06636},  
archivePrefix={arXiv},  
primaryClass={[cs.CL](http://cs.CL)},  
url={[https://arxiv.org/abs/2406.06636](https://arxiv.org/abs/2406.06636)},

}

@article{kroenke2009phq,
  title={The PHQ-8 as a measure of current depression in the general population},
  author={Kroenke, Kurt and Strine, Tara W and Spitzer, Robert L and Williams, Janet BW and Berry, Joyce T and Mokdad, Ali H},
  journal={Journal of affective disorders},
  volume={114},
  number={1-3},
  pages={163--173},
  year={2009},
  publisher={Elsevier}
}

@article{weathers1994ptsd,
  title={PTSD checklist—civilian version},
  author={Weathers, Frank W and Litz, B and Herman, D and Juska, J and Keane, T},
  journal={Journal of Occupational Health Psychology},
  year={1994}
}

@article{derogatis2010symptom,
  title={Symptom checklist-90-revised},
  author={Derogatis, Leonard R and Unger, Rachael},
  journal={The Corsini encyclopedia of psychology},
  pages={1--2},
  year={2010},
  publisher={Wiley Online Library}
}

@inproceedings{perez-rosas-etal-2019-makes,
    title = "What Makes a Good Counselor? Learning to Distinguish between High-quality and Low-quality Counseling Conversations",
    author = "P{\'e}rez-Rosas, Ver{\'o}nica  and
      Wu, Xinyi  and
      Resnicow, Kenneth  and
      Mihalcea, Rada",
    editor = "Korhonen, Anna  and
      Traum, David  and
      M{\`a}rquez, Llu{\'\i}s",
    booktitle = "Proceedings of the 57th Annual Meeting of the Association for Computational Linguistics",
    month = jul,
    year = "2019",
    address = "Florence, Italy",
    publisher = "Association for Computational Linguistics",
    url = "https://aclanthology.org/P19-1088",
    doi = "10.18653/v1/P19-1088",
    pages = "926--935",
}

@Article{fi15030110,
AUTHOR = {Wu, Zixiu and Balloccu, Simone and Kumar, Vivek and Helaoui, Rim and Reforgiato Recupero, Diego and Riboni, Daniele},
TITLE = {Creation, Analysis and Evaluation of AnnoMI, a Dataset of Expert-Annotated Counselling Dialogues},
JOURNAL = {Future Internet},
VOLUME = {15},
YEAR = {2023},
NUMBER = {3},
ARTICLE-NUMBER = {110},
URL = {https://www.mdpi.com/1999-5903/15/3/110},
ISSN = {1999-5903},
DOI = {10.3390/fi15030110}
}

@misc{luo2023chatgptfactualinconsistencyevaluator,
      title={ChatGPT as a Factual Inconsistency Evaluator for Text Summarization}, 
      author={Zheheng Luo and Qianqian Xie and Sophia Ananiadou},
      year={2023},
      eprint={2303.15621},
      archivePrefix={arXiv},
      primaryClass={cs.CL},
      url={https://arxiv.org/abs/2303.15621}, 
}

@misc{zhao2023surveylargelanguagemodels,
      title={A Survey of Large Language Models}, 
      author={Wayne Xin Zhao and Kun Zhou and Junyi Li and Tianyi Tang and Xiaolei Wang and Yupeng Hou and Yingqian Min and Beichen Zhang and Junjie Zhang and Zican Dong and Yifan Du and Chen Yang and Yushuo Chen and Zhipeng Chen and Jinhao Jiang and Ruiyang Ren and Yifan Li and Xinyu Tang and Zikang Liu and Peiyu Liu and Jian-Yun Nie and Ji-Rong Wen},
      year={2023},
      eprint={2303.18223},
      archivePrefix={arXiv},
      primaryClass={cs.CL},
      url={https://arxiv.org/abs/2303.18223}, 
}

@inproceedings{ji-etal-2022-mentalbert,
    title = "{M}ental{BERT}: Publicly Available Pretrained Language Models for Mental Healthcare",
    author = "Ji, Shaoxiong  and
      Zhang, Tianlin  and
      Ansari, Luna  and
      Fu, Jie  and
      Tiwari, Prayag  and
      Cambria, Erik",
    editor = "Calzolari, Nicoletta  and
      B{\'e}chet, Fr{\'e}d{\'e}ric  and
      Blache, Philippe  and
      Choukri, Khalid  and
      Cieri, Christopher  and
      Declerck, Thierry  and
      Goggi, Sara  and
      Isahara, Hitoshi  and
      Maegaard, Bente  and
      Mariani, Joseph  and
      Mazo, H{\'e}l{\`e}ne  and
      Odijk, Jan  and
      Piperidis, Stelios",
    booktitle = "Proceedings of the Thirteenth Language Resources and Evaluation Conference",
    month = jun,
    year = "2022",
    address = "Marseille, France",
    publisher = "European Language Resources Association",
    url = "https://aclanthology.org/2022.lrec-1.778",
    pages = "7184--7190",
}

@misc{zhai2024chinese,
      title={Chinese MentalBERT: Domain-Adaptive Pre-training on Social Media for Chinese Mental Health Text Analysis}, 
      author={Wei Zhai and Hongzhi Qi and Qing Zhao and Jianqiang Li and Ziqi Wang and Han Wang and Bing Xiang Yang and Guanghui Fu},
      year={2024},
      eprint={2402.09151},
      archivePrefix={arXiv},
      primaryClass={cs.CL}
}

@inproceedings{gratch-etal-2014-distress,
    title = "The Distress Analysis Interview Corpus of human and computer interviews",
    author = "Gratch, Jonathan  and
      Artstein, Ron  and
      Lucas, Gale  and
      Stratou, Giota  and
      Scherer, Stefan  and
      Nazarian, Angela  and
      Wood, Rachel  and
      Boberg, Jill  and
      DeVault, David  and
      Marsella, Stacy  and
      Traum, David  and
      Rizzo, Skip  and
      Morency, Louis-Philippe",
    editor = "Calzolari, Nicoletta  and
      Choukri, Khalid  and
      Declerck, Thierry  and
      Loftsson, Hrafn  and
      Maegaard, Bente  and
      Mariani, Joseph  and
      Moreno, Asuncion  and
      Odijk, Jan  and
      Piperidis, Stelios",
    booktitle = "Proceedings of the Ninth International Conference on Language Resources and Evaluation ({LREC}'14)",
    month = may,
    year = "2014",
    address = "Reykjavik, Iceland",
    publisher = "European Language Resources Association (ELRA)",
    url = "http://www.lrec-conf.org/proceedings/lrec2014/pdf/508_Paper.pdf",
    pages = "3123--3128",
}

@misc{chen2023llmempoweredchatbotspsychiatristpatient,
      title={LLM-empowered Chatbots for Psychiatrist and Patient Simulation: Application and Evaluation}, 
      author={Siyuan Chen and Mengyue Wu and Kenny Q. Zhu and Kunyao Lan and Zhiling Zhang and Lyuchun Cui},
      year={2023},
      eprint={2305.13614},
      archivePrefix={arXiv},
      primaryClass={cs.CL},
      url={https://arxiv.org/abs/2305.13614}, 
}

@misc{wang2024clientcenteredassessmentllmtherapists,
      title={Towards a Client-Centered Assessment of LLM Therapists by Client Simulation}, 
      author={Jiashuo Wang and Yang Xiao and Yanran Li and Changhe Song and Chunpu Xu and Chenhao Tan and Wenjie Li},
      year={2024},
      eprint={2406.12266},
      archivePrefix={arXiv},
      primaryClass={cs.CL},
      url={https://arxiv.org/abs/2406.12266}, 
}

@misc{dubey2024llama3herdmodels,
      title={The Llama 3 Herd of Models}, 
      author={Llama Team},
      year={2024},
      eprint={2407.21783},
      archivePrefix={arXiv},
      primaryClass={cs.AI},
      url={https://arxiv.org/abs/2407.21783}, 
}

@misc{yang2024qwen2technicalreport,
      title={Qwen2 Technical Report}, 
      author={An Yang and Baosong Yang and Binyuan Hui and Bo Zheng and Bowen Yu and Chang Zhou and Chengpeng Li and Chengyuan Li and Dayiheng Liu and Fei Huang and Guanting Dong and Haoran Wei and Huan Lin and Jialong Tang and Jialin Wang and Jian Yang and Jianhong Tu and Jianwei Zhang and Jianxin Ma and Jianxin Yang and Jin Xu and Jingren Zhou and Jinze Bai and Jinzheng He and Junyang Lin and Kai Dang and Keming Lu and Keqin Chen and Kexin Yang and Mei Li and Mingfeng Xue and Na Ni and Pei Zhang and Peng Wang and Ru Peng and Rui Men and Ruize Gao and Runji Lin and Shijie Wang and Shuai Bai and Sinan Tan and Tianhang Zhu and Tianhao Li and Tianyu Liu and Wenbin Ge and Xiaodong Deng and Xiaohuan Zhou and Xingzhang Ren and Xinyu Zhang and Xipin Wei and Xuancheng Ren and Xuejing Liu and Yang Fan and Yang Yao and Yichang Zhang and Yu Wan and Yunfei Chu and Yuqiong Liu and Zeyu Cui and Zhenru Zhang and Zhifang Guo and Zhihao Fan},
      year={2024},
      eprint={2407.10671},
      archivePrefix={arXiv},
      primaryClass={cs.CL},
      url={https://arxiv.org/abs/2407.10671}, 
}

@misc{jiang2024mixtralexperts,
      title={Mixtral of Experts}, 
      author={Albert Q. Jiang and Alexandre Sablayrolles and Antoine Roux and Arthur Mensch and Blanche Savary and Chris Bamford and Devendra Singh Chaplot and Diego de las Casas and Emma Bou Hanna and Florian Bressand and Gianna Lengyel and Guillaume Bour and Guillaume Lample and Lélio Renard Lavaud and Lucile Saulnier and Marie-Anne Lachaux and Pierre Stock and Sandeep Subramanian and Sophia Yang and Szymon Antoniak and Teven Le Scao and Théophile Gervet and Thibaut Lavril and Thomas Wang and Timothée Lacroix and William El Sayed},
      year={2024},
      eprint={2401.04088},
      archivePrefix={arXiv},
      primaryClass={cs.LG},
      url={https://arxiv.org/abs/2401.04088}, 
}

@misc{openai2024gpt4technicalreport,
      title={GPT-4 Technical Report}, 
      author={OpenAI Team},
      year={2024},
      eprint={2303.08774},
      archivePrefix={arXiv},
      primaryClass={cs.CL},
      url={https://arxiv.org/abs/2303.08774}, 
}

@misc{hua2024applying,
    title={Applying and Evaluating Large Language Models in Mental Health Care: A Scoping Review of Human-Assessed Generative Tasks},
    author={Yining Hua and Hongbin Na and Zehan Li and Fenglin Liu and Xiao Fang and David Clifton and John Torous},
    year={2024},
    eprint={2408.11288},
    archivePrefix={arXiv},
    primaryClass={cs.AI}
}

@article{johns2019systematic,
  title={A systematic review of therapist effects: A critical narrative update and refinement to review},
  author={Johns, Robert G and Barkham, Michael and Kellett, Stephen and Saxon, David},
  journal={Clinical Psychology Review},
  volume={67},
  pages={78--93},
  year={2019},
  publisher={Elsevier}
}

@inproceedings{zheng2023judging,
    title={Judging {LLM}-as-a-Judge with {MT}-Bench and Chatbot Arena},
    author={Lianmin Zheng and Wei-Lin Chiang and Ying Sheng and Siyuan Zhuang and Zhanghao Wu and Yonghao Zhuang and Zi Lin and Zhuohan Li and Dacheng Li and Eric Xing and Hao Zhang and Joseph E. Gonzalez and Ion Stoica},
    booktitle={Thirty-seventh Conference on Neural Information Processing Systems Datasets and Benchmarks Track},
    year={2023},
    url={https://openreview.net/forum?id=uccHPGDlao}
}

@inproceedings{liu-etal-2023-g,
    title = "{G}-Eval: {NLG} Evaluation using Gpt-4 with Better Human Alignment",
    author = "Liu, Yang  and
      Iter, Dan  and
      Xu, Yichong  and
      Wang, Shuohang  and
      Xu, Ruochen  and
      Zhu, Chenguang",
    editor = "Bouamor, Houda  and
      Pino, Juan  and
      Bali, Kalika",
    booktitle = "Proceedings of the 2023 Conference on Empirical Methods in Natural Language Processing",
    month = dec,
    year = "2023",
    address = "Singapore",
    publisher = "Association for Computational Linguistics",
    url = "https://aclanthology.org/2023.emnlp-main.153",
    doi = "10.18653/v1/2023.emnlp-main.153",
    pages = "2511--2522",
}

@inproceedings{wang-etal-2024-large-language-models-fair,
    title = "Large Language Models are not Fair Evaluators",
    author = "Wang, Peiyi  and
      Li, Lei  and
      Chen, Liang  and
      Cai, Zefan  and
      Zhu, Dawei  and
      Lin, Binghuai  and
      Cao, Yunbo  and
      Kong, Lingpeng  and
      Liu, Qi  and
      Liu, Tianyu  and
      Sui, Zhifang",
    editor = "Ku, Lun-Wei  and
      Martins, Andre  and
      Srikumar, Vivek",
    booktitle = "Proceedings of the 62nd Annual Meeting of the Association for Computational Linguistics (Volume 1: Long Papers)",
    month = aug,
    year = "2024",
    address = "Bangkok, Thailand",
    publisher = "Association for Computational Linguistics",
    url = "https://aclanthology.org/2024.acl-long.511",
    doi = "10.18653/v1/2024.acl-long.511",
    pages = "9440--9450",
}

@article{goldberg2020structure,
  title={The structure of competence: Evaluating the factor structure of the Cognitive Therapy Rating Scale},
  author={Goldberg, Simon B and Baldwin, Scott A and Merced, Kritzia and Caperton, Derek D and Imel, Zac E and Atkins, David C and Creed, Torrey},
  journal={Behavior Therapy},
  volume={51},
  number={1},
  pages={113--122},
  year={2020},
  publisher={Elsevier}
}

@article{hatcher2006development,
  title={Development and validation of a revised short version of the Working Alliance Inventory},
  author={Hatcher, Robert L and Gillaspy, J Arthur},
  journal={Psychotherapy research},
  volume={16},
  number={1},
  pages={12--25},
  year={2006},
  publisher={Taylor \& Francis}
}

@article{goodson2017impact,
  title={The impact of service-connected disability and therapist experience on outcomes from prolonged exposure therapy with veterans.},
  author={Goodson, Jason T and Helstrom, Amy W and Marino, Emily J and Smith, Rachel V},
  journal={Psychological Trauma: Theory, Research, Practice, and Policy},
  volume={9},
  number={6},
  pages={647},
  year={2017},
  publisher={Educational Publishing Foundation}
}

@article{mason2016my,
  title={My therapist is a student? The impact of therapist experience and client severity on cognitive behavioural therapy outcomes for people with anxiety disorders},
  author={Mason, Liam and Grey, Nick and Veale, David},
  journal={Behavioural and Cognitive Psychotherapy},
  volume={44},
  number={2},
  pages={193--202},
  year={2016},
  publisher={Cambridge University Press}
}

@article{robinson2009therapist,
  title={When therapist variables and the client's theory of change meet},
  author={Robinson, Bill},
  journal={Psychotherapy in Australia},
  volume={15},
  number={4},
  pages={60--65},
  year={2009},
  publisher={PsychOz Publications Kew, Vic.}
}

@article{spoto2013theoretical,
  title={Theoretical foundations and clinical implications of formal psychological assessment},
  author={Spoto, Andrea and Bottesi, Gioia and Sanavio, Ezio and Vidotto, Giulio},
  journal={Psychotherapy and psychosomatics},
  volume={82},
  number={3},
  pages={197--199},
  year={2013},
  publisher={S. Karger AG}
}

@article{meyer2001psychological,
  title={Psychological testing and psychological assessment: A review of evidence and issues.},
  author={Meyer, Gregory J and Finn, Stephen E and Eyde, Lorraine D and Kay, Gary G and Moreland, Kevin L and Dies, Robert R and Eisman, Elena J and Kubiszyn, Tom W and Reed, Geoffrey M},
  journal={American psychologist},
  volume={56},
  number={2},
  pages={128},
  year={2001},
  publisher={American Psychological Association}
}

@article{montazeri200312,
  title={The 12-item General Health Questionnaire (GHQ-12): translation and validation study of the Iranian version},
  author={Montazeri, Ali and Harirchi, Amir Mahmood and Shariati, Mohammad and Garmaroudi, Gholamreza and Ebadi, Mehdi and Fateh, Abolfazl},
  journal={Health and quality of life outcomes},
  volume={1},
  pages={1--4},
  year={2003},
  publisher={Springer}
}

@article{derogatis1983brief,
  title={The brief symptom inventory: an introductory report},
  author={Derogatis, Leonard R and Melisaratos, Nick},
  journal={Psychological medicine},
  volume={13},
  number={3},
  pages={595--605},
  year={1983},
  publisher={Cambridge University Press}
}

@misc{wen2024benchmarking,
    title={Benchmarking Complex Instruction-Following with Multiple Constraints Composition},
    author={Bosi Wen and Pei Ke and Xiaotao Gu and Lindong Wu and Hao Huang and Jinfeng Zhou and Wenchuang Li and Binxin Hu and Wendy Gao and Jiaxin Xu and Yiming Liu and Jie Tang and Hongning Wang and Minlie Huang},
    year={2024},
    eprint={2407.03978},
    archivePrefix={arXiv},
    primaryClass={cs.CL}
}

@article{hatfield2004use,
  title={The Use of Outcome Measures by Psychologists in Clinical Practice.},
  author={Hatfield, Derek R and Ogles, Benjamin M},
  journal={Professional Psychology: Research and Practice},
  volume={35},
  number={5},
  pages={485},
  year={2004},
  publisher={American Psychological Association}
}

@book{rogers2012client,
  title={Client centered therapy (new ed)},
  author={Rogers, Carl},
  year={2012},
  publisher={Hachette UK}
}

@article{hayes2020complex,
  title={A complex systems approach to the study of change in psychotherapy},
  author={Hayes, Adele M and Andrews, Leigh A},
  journal={BMC medicine},
  volume={18},
  pages={1--13},
  year={2020},
  publisher={Springer}
}

\section*{Language Resource References}
\label{lr:ref}
\bibliographystylelanguageresource{lrec2026-natbib}
\bibliographylanguageresource{languageresource}

\appendix
\section{Dataset Statistics}
\label{appendix:datasets}
As shown in Table~\ref{tab:datasets}, the datasets used in our study consist of both English and Chinese dialogue data. The High-Low Quality Counseling + AnnoMI dataset provides a challenging benchmark due to its long conversations, averaging around 80 utterances per session. In contrast, the TheraPhase dataset, which is constructed to simulate multi-session counseling, consists of shorter but more focused interactions, with an average of 11.5 utterances per session and a significantly higher word count per utterance. These structural differences highlight the diverse nature of our datasets, ensuring that models are tested under varying conversational conditions.

\begin{table*}[]
\centering
\resizebox{\textwidth}{!}{
\begin{tabular}{cccccc}
\toprule[1.25pt]
\textbf{Datasets} & \textbf{Language} & \textbf{\# of Clients} & \textbf{\# of Sessions} & \textbf{Avg. \# of Utterances} & \textbf{Words per Utterance} \\ \midrule
\makecell{High-Low Quality Counseling\\AnnoMI}        & English           & 110                   & 110                  & \makecell{79.8  \\(std = 26.1)} & \makecell{22.2\\(std = 27.1)} \\ \midrule
TheraPhase        & Chinese           & 400                   & 800                  & \makecell{11.5\\(std = 6.3)}  & \makecell{41.7\\(std = 20.9)} \\ 
\bottomrule
\end{tabular}
}
\vspace{-5pt}
\caption{\small Summary of key characteristics of the selected datasets, including language, number of clients, sessions, average number of utterances per session, and the average word count per utterance.}
\vspace{-5pt}
\label{tab:datasets}
\end{table*}

\section{Scoring Criteria for Symptom Assessment}
\label{appendix:symptom_assessment}
Table~\ref{tab:sa} provides an overview of the scoring system for symptom assessment. It is worth noting that, unlike traditional questionnaires where clients can provide imaginative responses to questions that are not directly addressed during a consultation, our evaluation process may leave certain symptoms unaddressed. Consequently, we include a score of -1 to indicate that the symptom was not addressed.

\begin{table*}[]
\centering
\resizebox{\textwidth}{!}{
\begin{tabular}{l|cccccc|cc}
\toprule[1.5pt]
\textbf{Models} & \textbf{Accuracy} $\uparrow$ & \textbf{Precision} $\uparrow$ & \textbf{Recall} $\uparrow$ & \textbf{F1}\textsubscript{Binary} $\uparrow$ & \textbf{F1}\textsubscript{Macro} $\uparrow$ & \textbf{F1}\textsubscript{Weighted} $\uparrow$ & \textbf{MSE} $\downarrow$ & \textbf{MAE} $\downarrow$ \\ \midrule
GPT-4          & 0.7744{\scriptsize$\pm$0.01}      & 0.6792{\scriptsize$\pm$0.01}      & 0.7187{\scriptsize$\pm$0.01}      & 0.6984{\scriptsize$\pm$0.01}      & 0.7591{\scriptsize$\pm$0.01}      & 0.7757{\scriptsize$\pm$0.01}      & 0.1369{\scriptsize$\pm$0.02}      & 0.2398{\scriptsize$\pm$0.01}      \\
GPT-4o         & \cellcolor{B}0.7833{\scriptsize$\pm$0.02}      & 0.6674{\scriptsize$\pm$0.02}      & 0.8043{\scriptsize$\pm$0.03}      & \cellcolor{B}0.7295{\scriptsize$\pm$0.02}      & \cellcolor{B}0.7744{\scriptsize$\pm$0.02}      & \cellcolor{B}0.7867{\scriptsize$\pm$0.02}      & \cellcolor{B}0.1207{\scriptsize$\pm$0.01}      & \cellcolor{B}0.2272{\scriptsize$\pm$0.02}      \\
GPT-4-turbo    & 0.7800{\scriptsize$\pm$0.01}      & \cellcolor{B}0.7503{\scriptsize$\pm$0.03}      & 0.5933{\scriptsize$\pm$0.01}      & 0.6623{\scriptsize$\pm$0.01}      & 0.7495{\scriptsize$\pm$0.01}      & 0.7734{\scriptsize$\pm$0.01}      & 0.2379{\scriptsize$\pm$0.03}      & 0.3754{\scriptsize$\pm$0.03}      \\
GPT-4o-mini    & 0.4844{\scriptsize$\pm$0.04}      & 0.4079{\scriptsize$\pm$0.02}      & \cellcolor{B}0.9144{\scriptsize$\pm$0.04}      & 0.5634{\scriptsize$\pm$0.01}      & 0.4641{\scriptsize$\pm$0.05}      & 0.4370{\scriptsize$\pm$0.06}      & 0.1962{\scriptsize$\pm$0.03}      & 0.3265{\scriptsize$\pm$0.02}      \\ 
\bottomrule
\end{tabular}
}
\vspace{-5pt}
\caption{\small Comparison of different models on various performance metrics using human-annotated data in psychological assessment. Metrics with an upward arrow $\uparrow$ indicate higher values are better, while metrics with a downward arrow $\downarrow$ indicate lower values are better. The results show mean values along with standard deviations for each metric. Cells highlighted in \colorbox{B}{blue} represent the best-performing results.}
\label{tab:RQ1H}
\vspace{-10pt}
\end{table*}
\begin{table*}[]
\centering
\resizebox{0.8\textwidth}{!}{
\begin{tabular}{l|cccccc}
\toprule[1.5pt]
\textbf{Models} & \textbf{Accuracy} $\uparrow$ & \textbf{Precision} $\uparrow$ & \textbf{Recall} $\uparrow$ & \textbf{F1}\textsubscript{Binary} $\uparrow$ & \textbf{F1}\textsubscript{Macro} $\uparrow$ & \textbf{F1}\textsubscript{Weighted} $\uparrow$ \\ \midrule
GPT-4          & \cellcolor{B}0.7444{\scriptsize$\pm$0.06}      & 0.8285{\scriptsize$\pm$0.04}      & \cellcolor{B}0.8406{\scriptsize$\pm$0.04}      & \cellcolor{B}0.8344{\scriptsize$\pm$0.04}      & 0.6370{\scriptsize$\pm$0.08}      & \cellcolor{B}0.7423{\scriptsize$\pm$0.06}      \\
GPT-4o         & 0.6778{\scriptsize$\pm$0.06}      & 0.8219{\scriptsize$\pm$0.02}      & 0.7391{\scriptsize$\pm$0.07}      & 0.7770{\scriptsize$\pm$0.05}      & 0.5939{\scriptsize$\pm$0.05}      & 0.6916{\scriptsize$\pm$0.05}      \\
GPT-4-turbo    & 0.6778{\scriptsize$\pm$0.06}      & 0.8046{\scriptsize$\pm$0.02}      & 0.7681{\scriptsize$\pm$0.11}      & 0.7815{\scriptsize$\pm$0.05}      & 0.5660{\scriptsize$\pm$0.04}      & 0.6809{\scriptsize$\pm$0.04}      \\
GPT-4o-mini    & 0.7222{\scriptsize$\pm$0.04}      & \cellcolor{B}0.8625{\scriptsize$\pm$0.06}      & 0.7681{\scriptsize$\pm$0.05}      & 0.8090{\scriptsize$\pm$0.03}      & \cellcolor{B}0.6410{\scriptsize$\pm$0.08}      & 0.7306{\scriptsize$\pm$0.05}      \\ 
\bottomrule
\end{tabular}
}
\vspace{-5pt}
\caption{\small Comparison of different models on various performance metrics using human-annotated data in treatment outcomes.}
\label{tab:RQ2H}
\vspace{-10pt}
\end{table*}

\section{Detailed Experimental Setup and Results for References Generation}
\label{appendix:reference}
In this appendix, we provide detailed experimental settings and results for the References Generation task, covering both the psychological assessment and treatment outcomes evaluation.

\subsection{Psychological Assessment}
For the psychological assessment task, we manually annotated 30 client sessions, achieving a Cohen's kappa of 0.73. Table~\ref{tab:RQ1H} summarizes the performance of the evaluated models on various metrics including Accuracy, Precision, Recall, Binary F1, Macro F1, Weighted F1, Mean Squared Error (MSE), and Mean Absolute Error (MAE). The best performing results for each metric are highlighted in blue. The results show that GPT-4o achieved the highest Accuracy (78.33\%) and Binary F1 score (72.95\%), and was therefore selected as the gold standard model for generating reference scores in this task.

\subsection{Treatment Outcomes Evaluation}
For the treatment outcomes evaluation, we annotated 60 sessions, obtaining a Cohen's kappa of 0.81. Table~\ref{tab:RQ2H} presents a detailed comparison of the models on Accuracy, Precision, Recall, Binary F1, Macro F1, and Weighted F1. The experimental results indicate that GPT-4 delivered the best performance with an Accuracy of 74.44\%, along with superior Recall and Binary F1 scores. Consequently, GPT-4 was chosen as the gold standard model for this evaluation.

Overall, these experiments provide a comprehensive comparison of the four models (GPT-4, GPT-4o, GPT-4-turbo, and GPT-4o-mini) across different tasks, thereby justifying the selection of the gold standard models for References Generation.

\clearpage

\section{Items-Aware Reasoning Prompts}
\label{appendix:promptIA}

\begin{analysisbox}[Prompt: Items-Aware Reasoning]
\textbf{Role:}\\
Imagine you are a skilled psychologist adept at identifying potential symptoms from interview. You can explain how these symptoms relate to the SCL-90 symptom checklist and specific items within it. \\

\textbf{Directives:}\\
Your task is to determine the presence or absence of symptoms from the Client's statements and provide detailed reasons for your assessment. Extract specific parts related to SCL-90 symptoms from the Client's statements. For each extracted part, indicate whether the symptom is present or not, and explain why this text is related to the SCL-90 symptom and specific item. If a symptom is mentioned but not present, extract that part and explain why the symptom is not present.SCL-90 is a psychological symptom assessment tool with 90 items, each evaluating different aspects of psychological distress.\\

\textbf{Additional Information:}\\
Symptom Checklist-90:\\
$<$Psychometric Test$>$\\
Presence of Symptoms: Extract the relevant part of the Client's statement. Indicate that the symptom is present. Explain why this text indicates the presence of the SCL-90 symptom and specify the item. Absence of Symptoms: Extract the part where the symptom is mentioned but not present. Indicate that the symptom is not present. Explain why this text does not indicate the presence of the SCL-90 symptom despite the mention.\\

\textbf{Output Formatting:}\\
$<$Format Instructions$>$\\

\textbf{Client Information:}\\
$<$Interview$>$\\

Please extract specific parts related to SCL-90 symptoms from the Client's statements. For each extracted part, indicate whether the symptom is present or not, and explain why this text is related to the SCL-90 symptom and specific item. If a symptom is mentioned but not present, extract that part and explain why the symptom is not present.
\end{analysisbox}

\clearpage
\section{Psychological Assessment Prompts}
\begin{analysisbox}[Prompt: Psychological Assessment]
\label{appendix:promptPA}
\textbf{Role:}\\
As a psychologist specializing in this evaluation task, based on the following interview and the extracted Symptom Checklist-90 (SCL-90) symptom-related content and explanation, provide a qualitative score (-1-2) for each symptom category. \\

\textbf{Score Criteria:}\\
Scoring criteria: -1 (Symptom not addressed in the interview), 0 (Symptom addressed in the interview, but no symptoms found, no signs of distress or dysfunction), 1 (Minimal symptoms, minor indications of distress but no significant dysfunction), 2 (Clear symptoms, clear indications of distress and significant dysfunction).\\

\textbf{Directives:}\\
Please note that this qualitative assessment is based on the state at the end of the interview. There may be noticeable symptoms during the interview, but these symptoms may become clarified or alleviated as the discussion progresses.\\

\textbf{Additional Information:}\\
$<$Psychometric Test$>$\\

\textbf{Output Formatting:}\\
$<$Format Instructions$>$\\

\textbf{Client Information:}\\
$<$Interview$>$\\
$<$Item-aware Reasoning Result$>$\\

Please extract specific parts related to SCL-90 symptoms from the Client's statements. For each extracted part, indicate whether the symptom is present or not, and explain why this text is related to the SCL-90 symptom and specific item. If a symptom is mentioned but not present, extract that part and explain why the symptom is not present.
\end{analysisbox}

\clearpage
\section{Items-Aware Reasoning Output}
\label{appendix:IAexample}
Table~\ref{tab:test} shows an example of the Items-Aware Reasoning output derived from a therapy session transcript. In this example, the system extracts a key client statement—specifically, the client's admission of consistently prioritizing others over themselves—and categorizes it under the symptom category ``Interpersonal Sensibility.'' The output further identifies a specific symptom (``Feeling others do not understand the client or are unsympathetic'') and marks its presence as ``Yes.'' An explanation is provided to clarify that the client's behavior might indicate feelings of being misunderstood or a lack of empathy from others. Additionally, an assessment score for Interpersonal Sensitivity is generated, demonstrating how the system quantifies this symptom. This example illustrates the system's capability to reason about session content and map client statements to relevant psychological constructs.

\begin{table*}[]
    \centering
    \begin{tabular}{p{\linewidth}}
        \toprule
        \textsc{Session:} \\
        Therapist:	So, thank you for coming in today.\\
        Client:	Yes.\\
        Therapist:	How are you feeling today?\\
        Client:	I feel great actually.\\
        Therapist:	Yeah? Good.\\
        Client:	Yeah.\\
        Therapist:	Good.\\
        Client:	I feel good.\\
        Therapist:	And so you did your clarifications, value clarifications-\\
        Client:	Yeah.\\
        Therapist:	-and what are your top five?\\
        Client:	Yes. It was a good, uh, experience for me. It was different. It was different than usual. There were several things that were different, and, uh, the number one value that I put was self-respect. And I-I don't even know if self-respect has ever been in my top five let alone my number one.\\
        Therapist:	Really?\\
        Client:	Yeah. And, um—\\
        Therapist:	Do you have any idea why that is?\\
        Client:	I do have an idea, I think, why that is. Um, I think that there's been a few things that have happened recently and something that really came to my awareness, when I visited with my family, is that \textcolor{pink}{\textbf{I have consistently through my whole life, probably, put other people first. And I have consistently, uh, almost not even considered myself in the equation.}} It was, uh, kind of sad in a way, at the time that I realized it. Uh, I didn't realize how severe it actually was, but I was kind of glad that I realized it because I feel like it's never too late to change-\\
        Therapist:	True.\\
        Client:	-and I feel like I can- I can, uh, respect and value myself just as much as I have other people. I know that's important. And I feel like when I do that, I'm a better person for other people as well.\\
        Therapist:	Mm-hmm. By not putting yourself on the back burner so much?\\
        ......\\
        \midrule
        \textsc{Items-Aware Reasoning Result:}\\
        Client Statement: \textcolor{pink}{\textbf{I have consistently through my whole life, probably, put other people first. And I have consistently, uh, almost not even considered myself in the equation.}}\\
        Symptom Category: Interpersonal Sensibility\\
        Specific Symptom: Feeling others do not understand the client or are unsympathetic.\\
        Presence: Yes\\
        Explanation: The client's statement indicates that they have been prioritizing others over themselves, which could be a sign of feeling misunderstood or not receiving empathy from others.\\
        ......\\
        \midrule
        \textsc{Assessment Score:}\\
        .......; Interpersonal Sensitivity: 1;......
        \\
        \bottomrule
    \end{tabular}
    \caption{\small Items-Aware Reasoning Output Example}
    \label{tab:test}
\end{table*}

\section{Output Formatting Errors} \label{app:errors}
In our two experiments, OpenAI series models produced no errors in output formatting, whereas open-source models encountered numerous issues. Specifically, the Figure~\ref{fig:error} below shows the error statistics for open-source models during the Assessment task, with the main issue being incorrect output that did not follow the Pydantic-defined JSON format.
\begin{figure*}[t]
    \centering
    \includegraphics[width=0.5\linewidth]{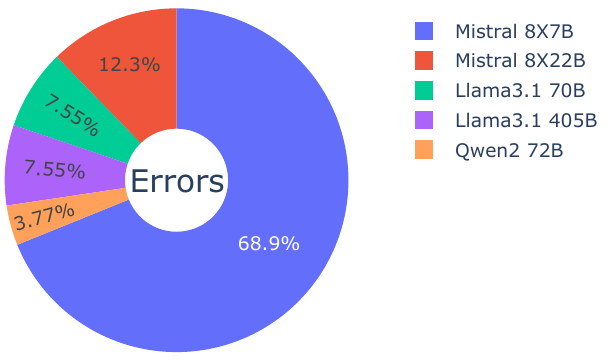}
    \vspace{-5pt}
    \caption{\small Error distribution across different models.}
    \label{fig:error}
\end{figure*}

\end{document}